\documentclass[letterpaper]{article} 
\usepackage{aaai2026}  
\usepackage{times}  
\usepackage{helvet}  
\usepackage{courier}  
\usepackage[hyphens]{url}  
\usepackage{graphicx} 
\usepackage{natbib}  
\usepackage{caption} 
\usepackage{algorithm}
\usepackage{algorithmic}
\usepackage[table]{xcolor}
\usepackage{makecell}
\usepackage{booktabs}
\usepackage{pifont} 

\newcommand{\cmark}{\ding{51}} 
\usepackage{amsmath}
\usepackage{amssymb} 
\usepackage{multirow}
\usepackage{lipsum}
\usepackage{xspace}
\makeatletter
\DeclareRobustCommand\onedot{\futurelet\@let@token\@onedot}
\def\@onedot{\ifx\@let@token.\else.\null\fi\xspace}

\def\etc{\emph{etc}\onedot}

\makeatother
\usepackage{newfloat}
\usepackage{listings}
\DeclareCaptionStyle{ruled}{labelfont=normalfont,labelsep=colon,strut=off} 
\floatstyle{ruled}
\newfloat{listing}{tb}{lst}{}
\floatname{listing}{Listing}
\title{Video Spatial Reasoning with Object-Centric 3D Rollout}
\author{
    Haoran Tang\textsuperscript{\rm 1$\dagger$}, 
    Meng Cao\textsuperscript{\rm 2,1$\dagger$},
    Ruyang Liu\textsuperscript{\rm 1},
    Xiaoxi Liang\textsuperscript{\rm 1},
    Linglong Li\textsuperscript{\rm 1},
    Ge Li\textsuperscript{\rm 1}\thanks{Corresponding Author},
    Xiaodan Liang\textsuperscript{\rm 3,2}
}
\affiliations{
    \textsuperscript{\rm 1} School of Electronic and Computer Engineering, Shenzhen Graduate School, Peking University \\
    \textsuperscript{\rm 2} Mohamed bin Zayed University of Artificial Intelligence \\
    \textsuperscript{\rm 3} Sun Yat-sen University \\
    \textsuperscript{$\dagger$}Equal Contribution

}

\usepackage{bibentry}

\begin{document}


\twocolumn[{
    \renewcommand\twocolumn[1][]{#1} 
    \maketitle
    \vspace{-1.0cm}
    \begin{center}
    \captionsetup{type=figure} 
    \includegraphics[width=1.0\textwidth]{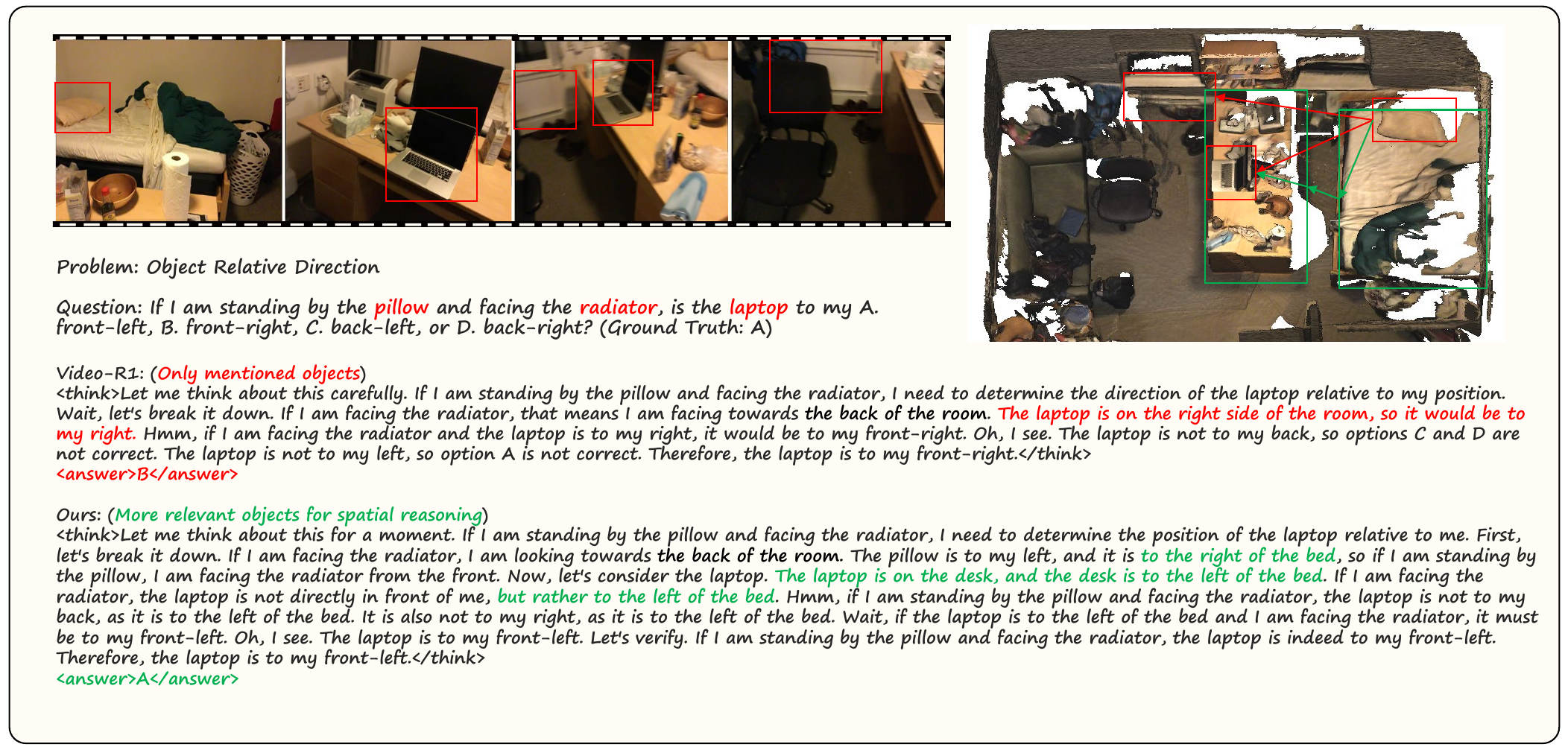} 
    \captionof{figure}{\textbf{Motivation of our proposed OCR.} The previous method exhibits \emph{query-locked reasoning}, focusing narrowly on objects explicitly mentioned in the prompt while ignoring critical contextual cues. Our method encourages MLLMs to rely on surrounding context and inter-object relationships for comprehensive spatial reasoning.}
    \vspace{0.5cm}
    \label{fig:motivation}
    \end{center}
}]

\begin{abstract}\label{abstract}
Recent advances in Multi-modal Large Language Models (MLLMs) have showcased remarkable capabilities in vision-language understanding. However, enabling robust video spatial reasoning—the ability to comprehend object locations, orientations, and inter-object relationships in dynamic 3D scenes—remains a key unsolved challenge. Existing approaches primarily rely on spatially grounded supervised fine-tuning or reinforcement learning, yet we observe that such models often exhibit \emph{query-locked reasoning}, focusing narrowly on objects explicitly mentioned in the prompt while ignoring critical contextual cues. To address this limitation, we propose Object-Centric 3D Rollout (OCR), a novel strategy that introduces structured perturbations to the 3D geometry of selected objects during training. By degrading object-specific visual cues and projecting the altered geometry into 2D space, OCR compels the model to reason holistically across the entire scene. We further design a rollout-based training pipeline that jointly leverages vanilla and region-noisy videos to optimize spatial reasoning trajectories. Experiments demonstrate state-of-the-art performance: our 3B-parameter model achieves 47.5\% accuracy on VSI-Bench, outperforming several 7B baselines. Ablations confirm OCR’s superiority over prior rollout strategies (e.g., T-GRPO, NoisyRollout).
\end{abstract}

\section{Introduction}\label{sec: introduction}

Human cognition effortlessly navigates and interprets the complex 3D spatial relationships inherent in dynamic visual scenes. We intuitively understand not just what objects are present, but where they are located relative to each other and how they are oriented in space. However, empowering Multi-modal Large Language Models (MLLMs) with analogous \emph{video spatial reasoning} \cite{chen2024spatialvlm,cheng2024spatialrgpt,cai2024spatialbot} capabilities remains a significant challenge, although they have achieved significant progress.

To enhance MLLMs' spatial reasoning capabilities, recent works have proposed constructing various large-scale datasets through spatially grounded Supervised Fine-Tuning (SFT) or Reinforcement Learning (RL). For instance, VSI-100K \cite{liao2025improved} and SpaceR-151K \cite{ouyang2025spacer} leverage ScanNet \cite{dai2017scannet} videos and 3D assets to generate synthetic yet richly annotated spatial reasoning tasks. While these datasets have led to performance gains on benchmark evaluations, our analysis reveals a critical limitation in the reasoning patterns learned by such models, \textbf{query-locked reasoning}. Specifically, MLLMs tend to focus narrowly on query-mentioned objects, neglecting the broader context of object-to-object spatial interactions. As illustrated in Figure \ref{fig:motivation}, when asked a question about the relative direction between objects, Video-R1~\cite{feng2025video} concentrates exclusively on the \texttt{pillow}, \texttt{laptop}, and \texttt{radiator}—objects directly mentioned in the query—while ignoring other spatially relevant items such as the \texttt{bed} and \texttt{desk}. These omitted entities provide important global cues necessary for accurate spatial inference, but are overlooked due to the model's over-reliance on query-salient tokens.

To overcome this limitation and encourage models to consider object-wise correlations, we propose an \textbf{O}bject-\textbf{C}entric 3D \textbf{R}ollout (\textbf{OCR}) strategy. The key idea is to perturb the spatial integrity of selected objects during training, thereby forcing MLLMs to reason through contextual relationships and leverage a more holistic understanding of the scene. Specifically, we randomly select 3D object assets from each video scene, inject structured noise into their 3D representations, and project the altered geometry back into the 2D space to generate region-noisy videos. Then we feed both the original (vanilla) and region-noisy videos into the GRPO rollout module to produce two parallel sets of reasoning trajectories. Both types of rollouts are aggregated to compute the final reward signals, encouraging the model to learn robust reasoning policies. This process selectively degrades visual cues associated with certain objects, compelling the model to rely on surrounding context and inter-object relationships to maintain accurate spatial reasoning. 

To facilitate our object-centric 3D rollout training, we construct a suite of SFT and RL datasets (dubbed as OCR-SFT and OCR-RL, respectively), which offers two key advantages over existing spatial reasoning datasets: 1) it encompasses a more diverse range of spatial perception types, and 2) it includes high-quality Chain-of-Thought (CoT) annotations as code-start data. Based on this dataset, our method achieves 47.5\% average results on the VSI-Bench benchmark at the cost of only 3B parameters, which is even higher than several existing 7B models. In addition, we conduct extensive ablation studies to demonstrate the superiority of our OCR to previous rollout strategies, including T-GRPO, NoisyRollout, \etc.

Our contributions are summarized as follows:
\begin{itemize}
    \item  We propose a novel object-centric rollout strategy that introduces structured noise to 3D object assets to enforce reasoning based on global scene context rather than memorizing salient tokens.

    \item A suite of SFT and RL datasets is proposed to facilitate our object-centric 3D rollout training with broader category coverage and high-quality CoT annotations.

    \item Extensive experimental and ablative results have demonstrated the superior performance of our proposed OCR training strategy. 
\end{itemize}

\section{Related Works}

\noindent \textbf{Video Spatial Reasoning.}
Recent advancements in large language models (LLMs)~\cite{hurst2024gpt, hui2024qwen2, guo2025deepseek} ~\nocite{tang2025muse, liu2024ppllava, liu2025flow4agent, cao2025video, cao2024rap, cao2024physgame, liu2024st} have motivated the development of multimodal LLMs (MLLMs) that extend reasoning capabilities to spatial understanding, particularly in embodied AI scenarios. Visual spatial reasoning has gained increasing attention, with models such as SpatialVLM~\cite{chen2024spatialvlm}, SpatialRGPT~\cite{cheng2024spatialrgpt}, and SpatialBot~\cite{cai2024spatialbot} utilizing grounding and depth estimation on 2D images to answer spatial questions. VSI-Bench~\cite{yang2025thinking} introduced the first benchmark for video-based spatial reasoning, facilitating evaluation on real-world video inputs. Building upon this setting, VG-LLM~\cite{zheng2025learning} and Spatial-MLLM~\cite{wu2025spatial} incorporate 3D scene reconstruction via VGGT~\cite{wang2025vggt} to extract spatial embeddings and apply supervised fine-tuning (SFT). In contrast to SFT-based methods, reinforcement learning (RL) approaches~\cite{feng2025video, ouyang2025spacer, liao2025improved}—particularly those leveraging GRPO~\cite{shao2024deepseekmath}—have shown superior adaptability and performance in spatial reasoning tasks. For example, Video-R1~\cite{feng2025video} introduces T-GRPO, employing shuffled video sequences for reward estimation, while SpaceR~\cite{ouyang2025spacer} incorporates 2D grid layouts as auxiliary supervision. Despite the progress, the reasoning process in 3D scenes always focuses on query-mentioned objects while neglecting the broader contexts. To bridge this gap, we propose an object-centric 3D rollout strategy that encourages more diverse and spatially grounded learning in complex environments by introducing noise to regions of the specific 3D objects.
~\nocite{cao2022locvtp, cao2021pursuit, cao2025ground, cao2024continual}


\noindent \textbf{GRPO Rollout Modification.}
Reinforcement learning has become a standard training paradigm for enhancing the reasoning capabilities of large language models (LLMs), as demonstrated in GPT-4o~\cite{hurst2024gpt}, Qwen-3~\cite{yang2025qwen3}, and DeepSeek-R1~\cite{guo2025deepseek}. Among various algorithms, Group Relative Policy Optimization (GRPO)~\cite{shao2024deepseekmath} has proven particularly effective in mathematical and code generation tasks by replacing the value model in PPO~\cite{schulman2017proximalpolicyoptimizationalgorithms} with group-based rollout. In the vision-language domain, GRPO has been adapted with modified rollout strategies. For instance, NoisyRollout~\cite{liu2025noisyrollout} injects visual noise into image inputs to improve the robustness of advantage estimation and encourage exploration of diverse reasoning paths. A series of methods \cite{cao2025ground,zheng2025deepeyes,zhang2025chain} incorporate image regional features into the CoT process, enabling the transformation of textual reasoning into visual-textual interleaved reasoning.

However, in the context of video spatial reasoning, such adaptations remain limited. Existing GRPO-based methods often struggle with overfitting due to the scarcity of diverse, high-quality 3D indoor video data, leading to repeated memorization of scene layouts. This highlights a critical gap and the need for task-specific rollout strategies that better accommodate the geometric and temporal complexity of video-based spatial reasoning tasks.

\begin{figure*}[t]
\centering
\includegraphics[width=1.0\textwidth]{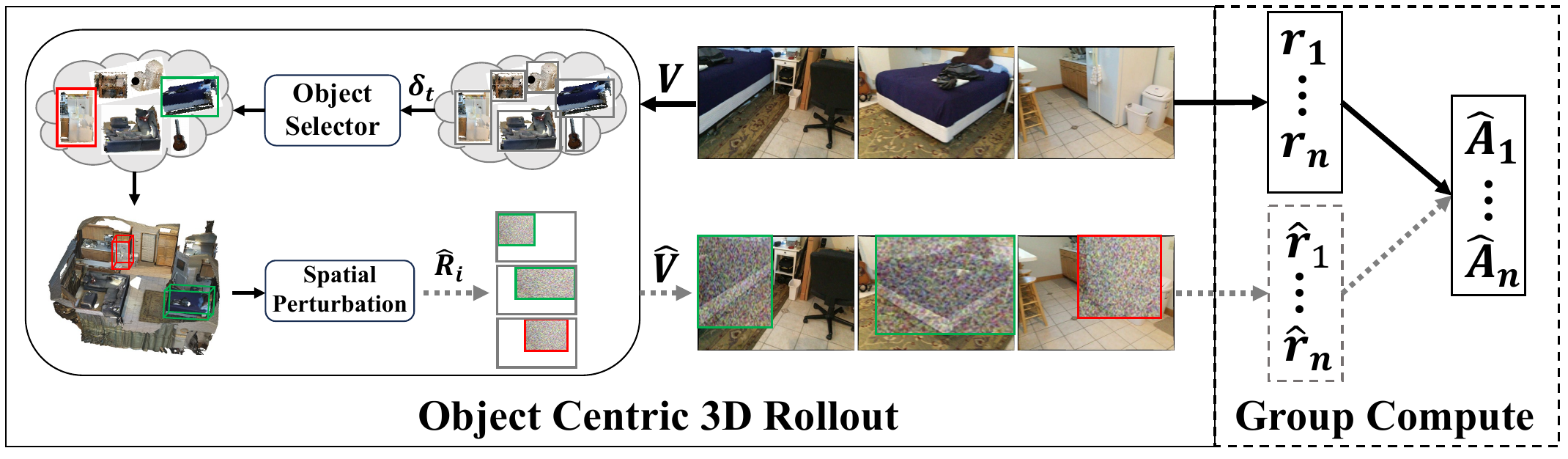}
\caption{\textbf{Overview of our Object-Centric 3D Rollout (OCR) strategy.} Given a video scene, we generate additional rollouts by injecting spatial noise into selected object regions using a step-wise object selector. These perturbed rollouts are used for advantage estimation, while only the clean rollouts (outlined in black) contribute to gradient updates.}
\label{fig:method}
\end{figure*}

\section{Methodology}\label{methods}

\subsection{Preliminary}
\noindent\textbf{Group Relative Policy Optimization. }
(GRPO)~\cite{shao2024deepseekmath} was originally developed and applied in DeepSeek-R1~\cite{guo2025deepseek} to enhance mathematical reasoning in natural language processing. However, it has also been effectively adapted to improve video-based spatial reasoning in recent advancements~\cite{feng2025video, ouyang2025spacer} with multimodal large language models (MLLMs). Given an input pair \( (V, \mathbf{q}) \) sampled from the training distribution, where \( V \) represents a video sequence derived from a 3D scene and \( \mathbf{q} \) denotes the corresponding textual query, the model generates outputs denoted as \( \mathbf{o} \). During the training stage, the model produces \( n \) response rollouts based on the previous policy \( \pi_{\theta_{\mathrm{old}}} \) denoted as \( \{(\mathbf{o}_i \mid V, \mathbf{q})\}_{i=1}^{n} \). A rule-based reward function \( r(V, \mathbf{q}, \mathbf{o}) \) is employed to prevent reward hacking. Specifically, the reward function assigns \( r(V, \mathbf{q}, \mathbf{o}) = 1 \) if the generated response \( \mathbf{o} \) is correct and adheres to the expected format. Otherwise, \( r(V, \mathbf{q}, \mathbf{o}) = 0 \). The rewards for all rollouts are represented as \( \mathbf{r} = \{ r_i \}_{i=1}^{n} = \{ r(V, \mathbf{q}, \mathbf{o}_i) \}_{i=1}^{n} \). GRPO computes the advantage of each rollout based on the mean reward of all generated rollouts, denoted as \( \mathrm{mean}(\mathbf{r}) \). The normalized advantage of the \( i \)-th rollout is then defined as:
\begin{equation}
    {A}_i = \frac{r_i - \mathrm{mean}(\mathbf{r})}{\mathrm{std}(\mathbf{r})}.
\end{equation}

Following Proximal Policy Optimization (PPO)~\cite{schulman2017proximalpolicyoptimizationalgorithms} formulation, the objective function of GRPO is defined as:
\begin{flalign}
\mathcal{J}_{\mathrm{GRPO}}(\theta) =  &\mathbb{E}_{(V, \mathbf{q}), \mathbf{o} \sim \pi_{\theta_{\mathrm{old}}}(\cdot \mid V, \mathbf{q})} \nonumber &\\
\Biggl[ \frac{1}{n} \sum_{i=1}^{n} \min \Biggl( &
\frac{\pi_{\theta}(\mathbf{o}_i \mid V, \mathbf{q})}{\pi_{\theta_{\mathrm{old}}}(\mathbf{o}_i \mid V, \mathbf{q})} A_i,\, \mathrm{clip} \Bigl( \frac{\pi_{\theta}(\mathbf{o}_i \mid V, \mathbf{q})}{\pi_{\theta_{\mathrm{old}}}(\mathbf{o}_i \mid V, \mathbf{q})}  , 1 - \epsilon, \nonumber &\\
1 + \epsilon \Bigr) A_i \Biggr )& - \beta \, \mathbb{D}_{\mathrm{KL}}(\pi_\theta \| \pi_{\mathrm{ref}}) \Big) \Bigg].&
\label{eq:grpo}
\end{flalign}
where $\pi_{\theta}$ is the current policy, $\epsilon>0$ sets the clipping range. 

\subsection{Object-Centric 3D Rollout}
While directly applying GRPO to video spatial reasoning has shown advantages over supervised fine-tuning (SFT), MLLMs still face significant limitations, as discussed in the Introduction. Specifically, it remains challenging for the model to perform complex spatial reasoning based solely on limited video content. To address this, we propose a method that incorporates an object-centric 3D rollout policy to enhance the GRPO training process. Our design includes object-guided spatial perturbation and step-wise annealed object selector.

\noindent\textbf{Object-Guided Spatial Perturbation.} \\
The key idea of object-aware spatial noisy injection is to create more challenging training examples, encouraging the model to reason about a broader set of objects beyond those mentioned in the prompt. To this end, we utilize the 3D bounding boxes of the input scene as augmented data. Specifically, for a scene $S$ represented by video $V$ containing $k$ input frames $\{f_{1},...,f_{k}\}$, we first extract the 3D bounding boxes of all visible objects, denoted as $\{\mathbf{B}_1,...,\mathbf{B}_M\}$. Each bounding box $\mathbf{B}_j$ consists of an axis-aligned 3D box defined by its center coordinates and dimensions. For each $\mathbf{B}_j$, we project all enclosed 3D points $(x, y, z) \in \mathbf{B}_j$ into 2D coordinates across all frames, forming the set $\{(x, y)_{f_1},..., (x, y)_{f_k}\}$. We denote this projected region across frames as $R_j = \{b_{j, f_1}, ..., b_{j, f_k}\}$,  where $b_{j, f_{k^{'}}}$ denotes the selected region of object $j$ on frame $k^{'}$. Given $m$ selected bounding boxes $\{\mathbf{B}_1, ..., \mathbf{B}_m\}$, we define the perturbed region of input frames as $\hat{R} = \bigcup_{j=1}^{m} R_j$. The resulting video input with noise is denoted as $\hat{V}$. For each input pair $(V, \mathbf{q})$, Gaussian noise is added to the region $\hat{R}$, forming a perturbed input pair $(\hat{V}, \mathbf{q})$.To increase GRPO training diversity, for every $n$ rollouts generated from $(V, \mathbf{q})$ using policy $\pi_{\theta_{\mathrm{old}}}$, we simultaneously generate $n$ spatially perturbed rollouts from $(\hat{V}, \mathbf{q})$, denoted as $\{(\hat{\mathbf{o}}_i \mid \hat{V}, \mathbf{q})\}_{i=n+1}^{2n}$, as illustrated in Figure \ref{fig:method}. The combined rewards are represented as:
\begin{equation}
     \hat{\mathbf{r}} = \{\hat{r}_i\}_{i=1}^{2n} = \{r(V, \mathbf{q}, \mathbf{o}_i)\}_{i=1}^{n} \cup \{r(\hat{V}, \mathbf{q}, \hat{\mathbf{o}}_i)\}_{i=n+1}^{2n} 
\end{equation}
The normalized advantage for each rollout is then calculated as:
\begin{equation}
    \hat{A}_i = \frac{\hat{r}_i - \mathrm{mean}(\hat{\mathbf{r}})}{\mathrm{std}(\hat{\mathbf{r}})}.
\end{equation}

\noindent\textbf{Step-wise Annealed Object Selector} \\
An essential question is how to choose which objects to perturb. In practice, we adopt a simple yet effective strategy: randomly selecting objects from the scene. However, since models typically learn from easy to hard cases during training, we propose a step-wise annealing strategy to encourage more complex spatial reasoning over time. This strategy involves gradually decreasing the number of objects used for perturbation, while simultaneously reducing the intensity of the applied Gaussian noise. We implement this using a scheduler $T_{\delta_t}(\cdot)$, where $\delta_t$ is a decay factor computed from the current training step $\alpha$ and total training steps $\beta$. Let $M$ be the total number of objects in scene $S$. We select $m'$ objects as:
\begin{equation}
    \{\mathbf{B}_j\}_{j=1}^{m'} = T_{\delta_t}(\{\mathbf{B}_j\}_{j=1}^{M}).
\end{equation}
In our experiments, $\delta_t$ is annealed linearly from 0.5 to 0 throughout training. Additional ablation studies are provided in ablative analysis.

Finally, the training objective of our proposed methods is represented as:
\begin{flalign}
\mathcal{J}_{\mathrm{GRPO}}(\theta) =  &\mathbb{E}_{(V, \mathbf{q}), \{\mathbf{o_{j}}\}_{1}^{n} \sim \pi_{\theta_{\mathrm{old}}}(\cdot \mid V, \mathbf{q}), \{\mathbf{\hat{o}_{j}}\}_{n+1}^{2n} \sim \pi_{\theta_{\mathrm{old}}}(\cdot \mid \hat{V}, \mathbf{q})}  \nonumber &\\
\Biggl[ \frac{1}{2n} \sum_{i=1}^{2n} \min \Biggl( &
\frac{\pi_{\theta}(\mathbf{o}_i \mid V, \mathbf{q})}{\pi_{\theta_{\mathrm{old}}}(\mathbf{o}_i \mid V, \mathbf{q})} \hat{A}_i,\, \mathrm{clip} \Bigl( \frac{\pi_{\theta}(\mathbf{o}_i \mid V, \mathbf{q})}{\pi_{\theta_{\mathrm{old}}}(\mathbf{o}_i \mid V, \mathbf{q})}  , 1 - \epsilon, \nonumber &\\
1 + \epsilon \Bigr) \hat{A}_i \Biggr )& - \beta \, \mathbb{D}_{\mathrm{KL}}(\pi_\theta \| \pi_{\mathrm{ref}}) \Big) \Bigg].&
\label{eq:grpo_noisy}
\end{flalign}

\noindent Note that only rollouts from the vanilla GRPO input pair are used to compute gradients and update the model parameters.

\section{Experiment}\label{experimental_results}
\begin{table}[h]
\centering
\caption{\textbf{Statistics of model-based data filtering.} This table summarizes the selection results and model configurations used to curate high-quality training samples.}

\begin{tabular}{|c|c|cc|c|}
    \toprule
    \multirow{2}{*}{\textbf{Num.}} & 
    \multirow{2}{*}{\textbf{Config}} & 
    \multicolumn{2}{c|}{\textbf{Training Data}} & 
    \multirow{2}{*}{\textbf{Acc. (\%)}} \\
    & & \textbf{Correct} & \textbf{Wrong} & \\
    \midrule
    \ding{172} & $\mathcal{F}^2$ &  24,391 & 73,712  & 24.86 \\
    \ding{173} & $\mathcal{F}^{16}$ &  25,287 & 72,816 & 25.78 \\
    \ding{174} & $(\mathcal{F}^{16})_{bev}$ & 26,380 & 71,723  & 26.89 \\
    \ding{175} & $(\mathcal{F}^{16})_{grpo}$ & 41,485 & 56,618 & 42.29 \\
    \bottomrule
\end{tabular}
\label{tab:data_construction}
\end{table}

\begin{table*}[t]
\renewcommand\arraystretch{1.1}
    \centering
    \caption{\textbf{Comparison with state-of-the-art models on VSI-Bench.} \textit{Spatial-SFT Models} are trained using supervised fine-tuning, while \textit{Spatial-RL Models} are trained using reinforcement learning approaches.
}
    \resizebox{\linewidth}{!}{
    \begin{tabular}{r|c|c|cccccccc}
    & & &
    \rotatebox{30}{\textbf{Obj. Count}} &
    \rotatebox{30}{\textbf{Abs. Dist.}} &
    \rotatebox{30}{\textbf{Obj. Size}} & 
    \rotatebox{30}{\textbf{Room Size}} &
    \rotatebox{30}{\textbf{Rel. Dist.}} &
    \rotatebox{30}{\textbf{Rel. Dir.}} &
    \rotatebox{30}{\textbf{Route Plan}} &
    \rotatebox{30}{\textbf{Appr. Order}} \\
    \makecell[c]{\textbf{Model}} & \textbf{\#Params} & \textbf{Avg.} & \multicolumn{4}{c}{\textbf{Numerical Answer}} & \multicolumn{4}{c}{\textbf{Multiple-Choice Answer}} \\
    \hline
    \rowcolor{lightgray!5}
    \hline
    
    \rowcolor{lightgray!5}
    \multicolumn{1}{l|}{\textcolor{black}{\textit{Proprietary Models (API)}}} & & & & & & & & & & \\
    GPT-4o & - & 34.0 & 46.2 & 5.3 & 43.8 & 38.2 & 37.0 & 41.3 & 31.5 & 28.5 \\
    Gemini-1.5-Flash & - & 42.1 & 49.8 & 30.8 & 53.5 & {54.4} & 37.7 & 41.0 & 31.5 & 37.8 \\
    Gemini-1.5-Pro & - & 45.4 & {56.2} & {30.9} & {64.1} & 43.6 & {51.3} & {46.3} & {36.0} & 34.6 \\
    \hline
    \rowcolor{lightgray!5}
    \multicolumn{1}{l|}{\textcolor{black}{\textit{Open-source Models}}} & & & & & & & & & & \\
    InternVL2-8B & 8B & 34.6 & 23.1 & {28.7} & 48.2 & {39.8} & 36.7 & 30.7 & 29.9 & 39.6 \\
    InternVL2-40B & 40B & 36.0 & 34.9 & 26.9 & 46.5 & 31.8 & 42.1 & 32.2 & 34.0 & 39.6 \\
    LLaVA-NeXT-Video-7B & 7B & 35.6 & 48.5 & 14.0 & 47.8 & 24.2 & {43.5} & 42.4 & 34.0 & 30.6 \\
    LLaVA-NeXT-Video-72B & 72B & 40.9 & {48.9} & 22.8 & 57.4 & 35.3 & 42.4 & 36.7 & {35.0} & {48.6} \\
    LLaVA-OneVision-7B & 7B & 32.4 & 47.7 & 20.2 & 47.4 & 12.3 & 42.5 & 35.2 & 29.4 & 24.4 \\
    LLaVA-OneVision-72B & 72B & 40.2 & 43.5 & 23.9 & {57.6} & 37.5 & 42.5 & 39.9 & 32.5 & 44.6 \\
    Qwen2.5-VL-7B-Instruct & 7B & 33.0 & 40.9 & 14.8 & 43.4 & 10.7 & 38.6 & 38.5  & 33.0 & 29.8 \\
    Qwen2.5-VL-72B-Instruct & 72B & 37.0 & 25.1 & 29.3 & 54.5 & 38.8 & 38.2 & 37.0  & 34.0 & 28.9 \\
    \hline
    \rowcolor{lightgray!5}
    \multicolumn{1}{l|}{\textcolor{black}{\textit{Spatial-SFT Models}}} & & & & & & & & & & \\
    SPAR & 8B & 41.1 & - & - & - & - & - & - & -  & - \\
    SpaceR(SFT) & 7B & 41.6 & - & - & - & - & - & - & -  & - \\
    VG-LLM & 4B & 46.1 & 66.4 & 36.6 & 55.2 & 56.3 & 40.8 & 43.4 & 30.4 & 39.5\\

    \hline
    \rowcolor{lightgray!5}
    \multicolumn{1}{l|}{\textcolor{black}{\textit{Spatial-RL Models}}} & & & & & & & & & & \\
    Video-R1 & 7B & 37.1 & - & - & - & - & - & - & -  & - \\
    vsGRPO-V & 7B & 40.7 & 59.9 & 29.6 & 50.8 & \textbf{48.3} & 35.4 & 35.6 & 34.0 & 31.5 \\
    SpaceR(SG-RLVR) & 7B & 45.6 & - & - & - & - & - & - & -  & - \\
    \textbf{Ours} & 3B & \textbf{47.5} & \textbf{63.2} & \textbf{34.1} & \textbf{57.4} & 46.7 & \textbf{39.6} & \textbf{45.5} & \textbf{44.3} & \textbf{49.8}\\
    
    \hline
    \end{tabular}
    }

\label{tab:vsibench}
\end{table*}

\begin{figure}[t]
\centering
\includegraphics[width=1.0\linewidth]{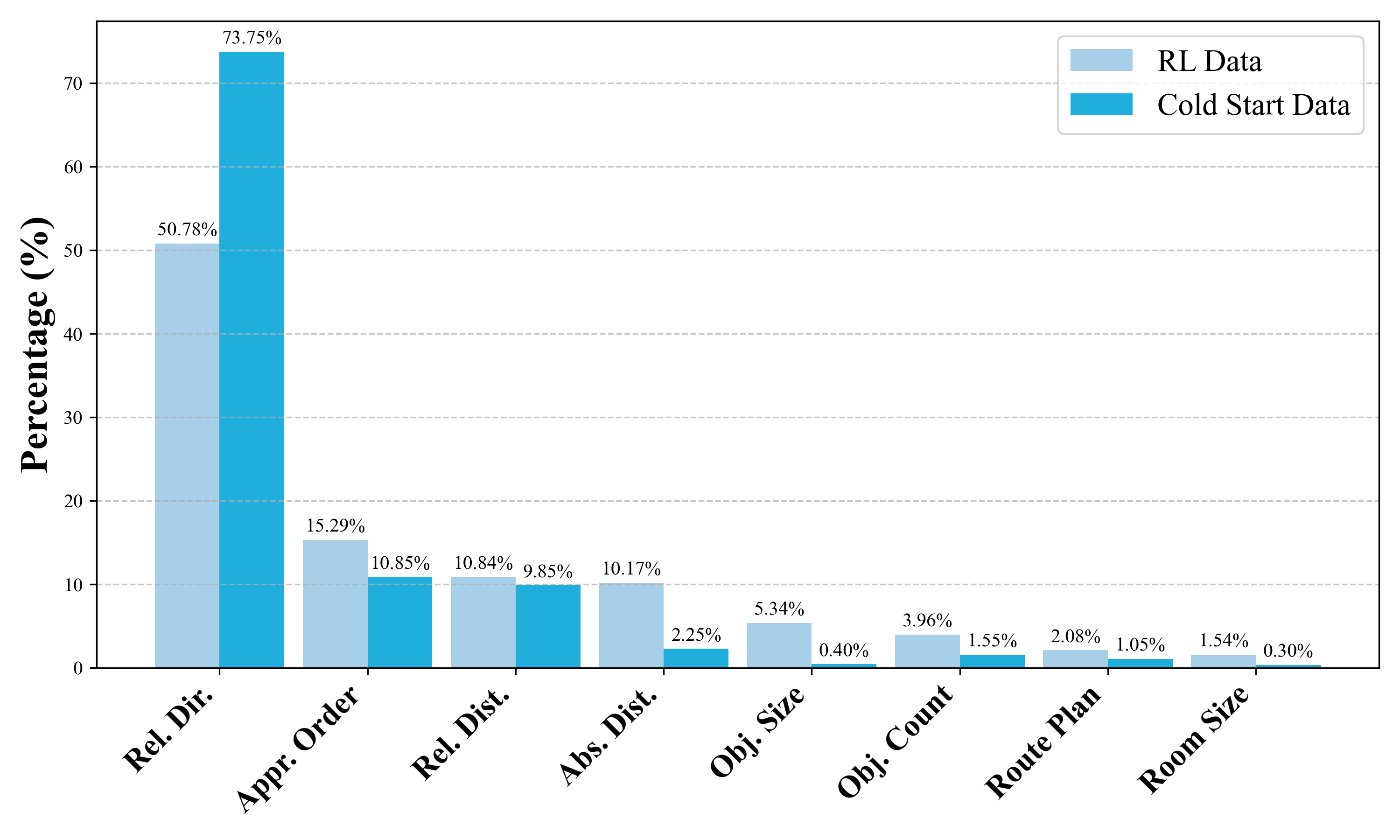}
\caption{\textbf{Distribution of the constructed dataset.} The chart illustrates the category-wise distribution of our data. Light blue denotes the training set, while blue indicates the curated cold-start subset. Categories are sorted in descending order based on their frequency in the RL data.}
\label{fig:data_distribution}
\end{figure}

\begin{table}[t]
\centering
\caption{\textbf{Ablations on the main component of our method.}}
\resizebox{1.0\linewidth}{!}{
\begin{tabular}{c|cc|cc|c}
    \toprule
    \textbf{Type} & \multicolumn{2}{c|}{\textbf{ColdStart Data}} & \multicolumn{2}{c|}{\textbf{Rollout Method}} & \textbf{VSI-Bench} \\
    & Video-R1 & Ours & Vanilla GRPO & Ours & Avg. \\
    \midrule
    1 & \cmark &  &  &  & 35.1 \\
    2 &  &  \cmark &  &  & 40.4 \\
    \hline
    3 & \cmark &  & \cmark &  & 42.6 \\
    4 &  & \cmark & \cmark &  & 43.9\\
    5 & \cmark &  &  & \cmark & 45.6 \\
    6 &  & \cmark &  & \cmark & \textbf{47.5} \\
    \bottomrule
\end{tabular}
}
\label{tab:ablate_components}
\end{table}

\begin{table}[t]
  \centering
  \caption{\textbf{Ablations on rollout policy.}}
  \resizebox{0.9\linewidth}{!}{
  \begin{tabular}{lccc}
    \toprule
    \textbf{Method} &  \textbf{VSI-Bench (Avg.)} \\
    \midrule
    Baseline        & 35.1 \\
    + Vanilla GRPO    & 43.9 \\
    + T-GRPO    & 44.1 \\
    + Noisy Rollout & 45.6 \\
    + Downsample Rollout & 44.2 \\
    + \textbf{Ours}       & \textbf{47.5} \\
    \bottomrule
  \end{tabular}
  }
\label{tab:ablate_noisy_policy}
\end{table}

\begin{table}[t]
  \centering
  \caption{
  \textbf{Ablations on object schedule.}}
  \resizebox{0.6\linewidth}{!}{
  \begin{tabular}{cccc}
    \toprule
    \textbf{Schedule} &  \textbf{VSI-Bench (Avg.)} \\
    \midrule
    \textit{Fix}  &  45.0 \\
    \textit{Linear}  & \textbf{47.5} \\
    \textit{Exp}    & 46.0 \\
    \textit{Cos}    & 46.3 \\
    \bottomrule
  \end{tabular}
  }
\label{tab:ablate_noisy_schedule}
\end{table}


\subsection{Data Construction}
As video spatial reasoning lacks high-quality training datasets in VSI-Bench~\cite{yang2025thinking}, we construct a comprehensive training dataset containing 98K samples, covering all video spatial categories listed in VSI-Bench. We name this dataset \textbf{OCR-RL}. Additionally, to enhance the model's spatial reasoning ability, we curate a smaller, high-quality subset with detailed spatial reasoning annotations for the cold start SFT stage, referred to as \textbf{OCR-SFT}. Details of the construction process are provided below.\\

\noindent \textbf{Training Data Construction.} For the training dataset, we follow the benchmark data generation practices outlined in VSI-Bench, excluding the categories of \textit{route planning} and \textit{appearance order}, as these require extensive manual annotation. To construct the dataset, for these two categories, we use the annotated data from SpaceR~\cite{ouyang2025spacer} and VLM-3R~\cite{fan2025vlm}. Low-quality annotations are filtered using GPT-4o~\cite{hurst2024gpt}, resulting in a final training set of 98K samples. The data distribution of our training dataset is shown in Figure~\ref{fig:data_distribution}. Compared with VSI-100K and SpaceR-151K, our dataset covers all categories in VSI-Bench, making the training process unbiased.\\
\noindent \textbf{Cold Start Data Construction.} The necessity of a cold start dataset arises from two key challenges: (1) the model must learn the reasoning and thinking format, and (2) modern models cannot acquire high-quality spatial reasoning capabilities solely through GRPO rollout policies. To address these challenges, we extract a subset from the training set that highlights the spatial reasoning deficiencies of the model. This subset is then used to generate high-quality spatial chain-of-thought (CoT) annotations using proprietary models such as GPT-4o~\cite{hurst2024gpt}. To identify hard spatial reasoning questions, we employ Qwen2.5-VL-7B-Instruct~\cite{bai2025qwen2} as a filtering model. The model is evaluated on the training set under four different settings, as shown in Table~\ref{tab:data_construction}. The settings include: \ding{172} 2-frame input; \ding{173} 16-frame input; \ding{174} 16-frame input with a global BEV (bird's-eye view) map; \ding{175} 16-frame input with the model fine-tuned for 2000 steps using the GRPO algorithm.

The accuracy results in Table~\ref{tab:data_construction} demonstrate that increasing the number of frames improves performance. Additionally, the inclusion of a BEV map provides global spatial information about the input scene, highlighting that modern models struggle with local spatial reasoning. Although GRPO training improves the model's ability to answer questions correctly, a significant portion of errors still stem from improper spatial reasoning. To construct the cold start dataset, we focus on hard spatial problems by filtering samples based on the following criteria: (1) Questions where the model transitions from incorrect to correct answers when provided with more frames (\ding{173} - \ding{172}); (2) Questions where the model transitions from incorrect to correct answers when provided with a global BEV map (\ding{174} - \ding{172}); (3) Questions that remain incorrectly answered even after GRPO training ($\neg$\ding{175}). To collect representative hard spatial problems, we filter samples from the intersections (\ding{173} - \ding{172}) $\land$ $\neg$\ding{175} and (\ding{174} - \ding{172}) $\land$ $\neg$\ding{175}, selecting 1000 samples for each category. For annotation generation, as shown in Table~\ref{tab:data_construction}, the BEV map, which includes global spatial information, proves useful for generating high-quality data. Therefore, we annotate the dataset using a BEV map extracted by ~\cite{wang2025ross3d} as the first image and 32 frames as the video input for GPT-4o annotations.

Figure~\ref{fig:data_distribution} illustrates the data distribution of our RL dataset and Cold Start SFT dataset. It can be observed that, through our data filtering process, more questions are selected from the categories of \textit{relative direction} and \textit{relative distance}, while fewer questions are selected from \textit{object size}, \textit{object count}, and \textit{room size}, as these are not the core challenges of hard spatial reasoning.

\subsection{Experimental Settings}
\noindent \textbf{Implementation Details.}  
All experiments are conducted on the VSI-Bench~\cite{yang2025thinking}, which comprises 5130 testing samples curated from ScanNet~\cite{dai2017scannet}, ScanNet++~\cite{yeshwanth2023scannet++}, and ARKitScenes~\cite{baruch2021arkitscenes}. We evaluate our approach using the Qwen2.5-VL-3B-Instruct~\cite{bai2025qwen2} model, with 16-frame inputs for training and 32-frame inputs for inference. For model training, we first perform supervised fine-tuning (SFT) on OCR-SFT for 2 epochs. Subsequently, GRPO training is conducted with 2000 steps on OCR-RL. The number of clean rollouts and perturbed rollouts is set to 4 and 4. Other configurations are set following Video-R1~\cite{feng2025video}. All experiments are executed on a cluster of 8 NVIDIA A100 GPUs.

\subsection{Performance Comparison}
We compare our Qwen2.5-VL-3B model trained with the proposed Object-Centric 3D Rollout (OCR) strategy against recent state-of-the-art models on the VSI-Bench. As shown in Table~\ref{tab:vsibench}, our 3B model achieves an average score of 47.5, surpassing several larger-scale models, including proprietary systems such as GPT-4o~\cite{hurst2024gpt} and Gemini-1.5-Pro~\cite{comanici2025gemini}, as well as open-source models like LLaVA-Next-Video-72B~\cite{zhang2024video}, LLaVA-OneVision~\cite{li2024llava}, and Qwen2.5-VL-72B~\cite{bai2025qwen2}. This demonstrates the strong effectiveness of our OCR training pipeline and high-quality curated dataset. To further validate the impact of our rollout strategy, we compare against both \textit{Spatial-SFT} and \textit{Spatial-RL} baselines. Among the \textit{Spatial-SFT} models~\cite{zhang2025flatland, ouyang2025spacer, zheng2025learning}, VG-LLM, built on the same Qwen2.5-VL-3B backbone, shows competitive performance—especially in object counting and absolute distance reasoning—benefiting from the powerful 3D encoding capabilities of VGGT~\cite{wang2025vggt}. However, this comes at the cost of increased computational overhead and slower inference. In the \textit{Spatial-RL} category~\cite{zheng2025video, liao2025improved, ouyang2025spacer}, our method consistently outperforms previous 7B-scale models that employ GRPO variants, such as Video-R1, vsGRPO-V, and SpaceR, highlighting the unique advantages of our object-centric rollout design. These results collectively confirm that incorporating structured object-level perturbations during training significantly enhances a model’s ability to reason spatially in complex 3D environments.

\subsection{Ablation Analysis}

We conduct comprehensive ablation experiments to analyze the contributions of each component in our proposed Object-Centric 3D Rollout (OCR) framework. As shown in Table~\ref{tab:ablate_components}, we first compare our curated cold-start dataset (OCR-SFT) against the one adopted in Video-R1~\cite{feng2025video}. Comparing setups (1 vs. 2), (3 vs. 4), and (5 vs. 6), we observe consistent improvements when using our dataset, both in the initial cold-start phase and during full GRPO training. These results highlight two key benefits: (1) our annotated data offers high-quality Chain-of-Thought(CoT) guidance, directly supporting spatial reasoning, and (2) the model-based filtering strategy effectively selects challenging samples that require deeper reasoning, thereby enriching the training distribution. To isolate the effect of our proposed rollout mechanism, we compare vanilla GRPO rollouts with our object-centric 3D rollout (comparing 3 vs. 5 and 4 vs. 6). The consistent performance gain across these pairs demonstrates the effectiveness of injecting structured spatial perturbations into the rollout policy.

Further analysis of rollout variants is presented in Table~\ref{tab:ablate_noisy_policy}. We benchmark our method against T-GRPO~\cite{feng2025video}, NoisyRollout~\cite{liu2025noisyrollout}, and a simple frame-downsampling strategy. For NoisyRollout, we extend image-based noise augmentation to all video frames. The downsampling variant randomly removes frames following the same scheduling function used in OCR. While NoisyRollout improves robustness in spatial tasks, our object-centric perturbation achieves superior performance by explicitly encoding spatial object interactions.

Finally, we evaluate different scheduling strategies used in the object selector. We evaluate four perturbation schedules: \texttt{Fix} (a constant 25\% object perturbation), \texttt{Linear} (linearly decreasing from 50\% to 0\%), \texttt{Exp}, and \texttt{Cos} (both using exponential and cosine decay, respectively, with the same range as \texttt{Linear}). We find that the \texttt{Linear} schedule yields the best performance under GRPO training. This schedule encourages the model to learn from easier cases initially and gradually adapt to harder scenarios, thus discovering more diverse reasoning paths. All experiments are conducted using the Qwen2.5-VL-3B~\cite{bai2025qwen2} backbone to ensure consistent comparisons.

\begin{figure*}[!ht]
\centering
\includegraphics[width=0.9\textwidth]{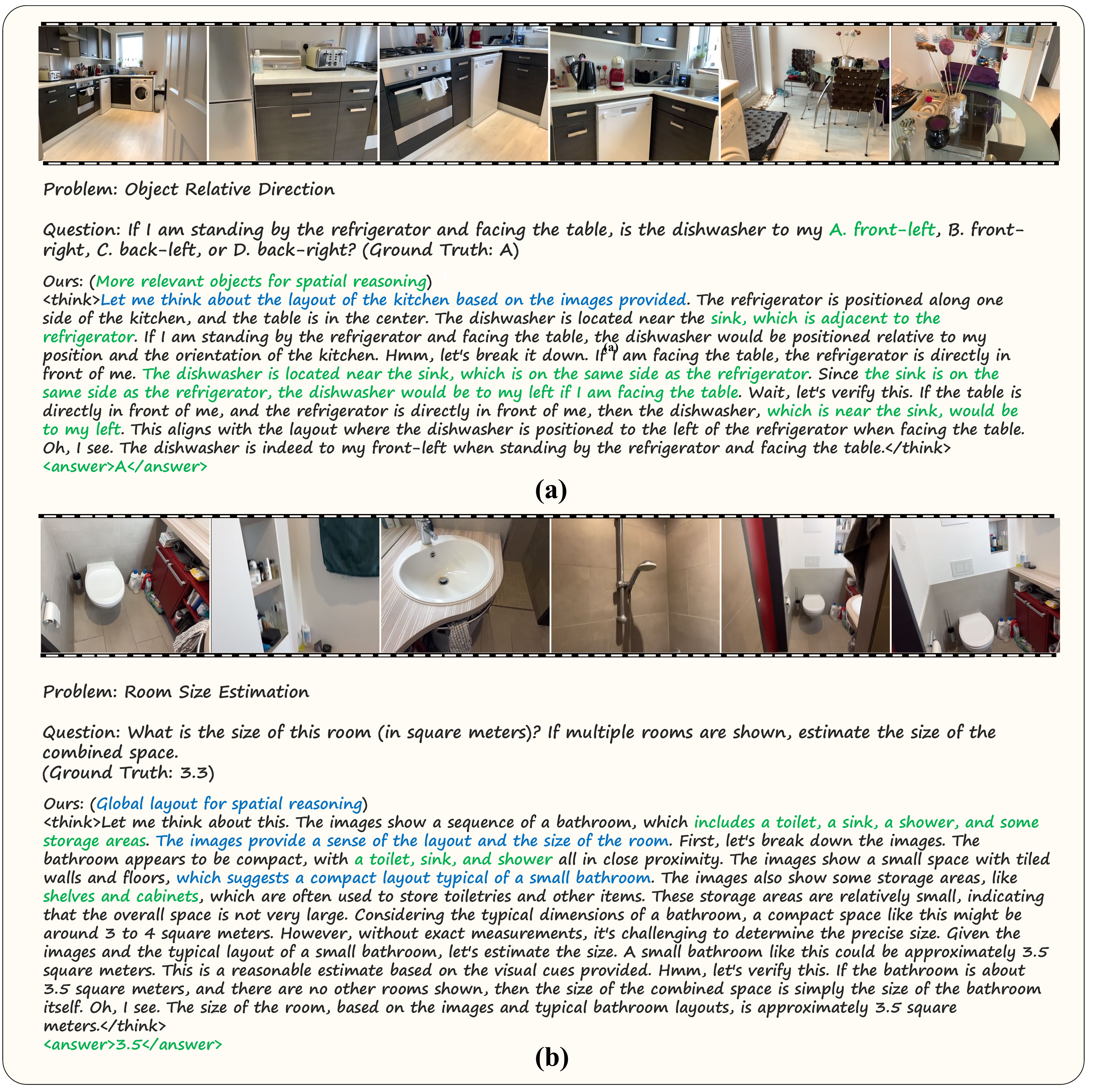}
\caption{\textbf{Visualization samples of our method.} Text in green indicates instances where the model incorporates additional relevant objects beyond those mentioned in the query, enhancing spatial reasoning. Text in blue reflects the model's ability to reason from a global scene perspective rather than relying solely on local visual cues.}
\label{fig:visualization}
\end{figure*}

\subsection{Visualization Results}

We provide additional qualitative results on VSI-Bench in Figure~\ref{fig:visualization} to demonstrate the effectiveness of our proposed GRPO training strategy. The visualizations reveal two key advantages of our method. \textbf{(1) Reasoning with Extra Objects.} 
As highlighted in green in Figure~\ref{fig:visualization}(a) and (b), our model frequently leverages additional objects not explicitly mentioned in the question. In Figure~\ref{fig:visualization}(a), the model infers the relative positions of the \texttt{refrigerator}, \texttt{table}, and \texttt{dishwasher} by incorporating the \textbf{sink}, which provides essential spatial cues absent from the original query and frames. Similarly, in Figure~\ref{fig:visualization}(b), the model utilizes contextual objects such as the \texttt{shower}, \texttt{shelves}, and \texttt{cabinets} to identify the room type and estimate its size based on object arrangement. These examples demonstrate the model's ability to introduce relevant auxiliary objects to enhance spatial reasoning. \textbf{(2) Global Spatial Perspective.} 
While existing spatial reasoning models tend to rely on local visual cues, our model exhibits a broader understanding by first establishing a global layout of the scene before addressing the specific question. In Figure~\ref{fig:visualization}(a), the model outlines a holistic scene layout and then selects the \textbf{sink} to support finer-grained spatial inference. In Figure~\ref{fig:visualization}(b), the model first composes a global view of the bathroom and then accurately estimates its size using the observed object configuration. These findings further demonstrate that our object-centric 3D rollout policy encourages the model to reason beyond explicitly mentioned entities and promotes spatially coherent reasoning grounded in 3D scene structure.

\section{Conclusion}\label{sec:conclusion}

We present Object-Centric 3D Rollout (OCR), a novel training strategy for enhancing video spatial reasoning in MLLMs. By introducing structured 3D perturbations to object regions and jointly leveraging vanilla and region-noisy rollouts, OCR encourages holistic scene understanding beyond query-mentioned objects. Trained on a curated dataset with diverse spatial scenarios and high-quality CoT annotations, our 3B model achieves state-of-the-art results on VSI-Bench, outperforming several 7B baselines. This work underscores the value of geometry-aware rollouts and paves the way for more context-grounded spatial reasoning in multimodal systems.

\section{Acknowledgments}
This work is supported by the National Key Research and Development Program of China (2024YFE0203100).

\bibliography{aaai2026}
\end{document}